%% file: main.tex
\documentclass[sigconf]{acmart}
\usepackage{subfigure}
\usepackage{amssymb}

\def\BibTeX{{\rm B\kern-.05em{\sc i\kern-.025em b}\kern-.08emT\kern-.1667em\lower.7ex\hbox{E}\kern-.125emX}}
    
\copyrightyear{2019}
\acmYear{2019}
\setcopyright{acmlicensed}
\acmConference[AutoML'19]{The Third International Workshop on Automation in Machine Learning and Big Data}{August 05, 2019}{Anchorage, AK}
 \acmPrice{15.00}

\begin{document}

%
\title[Augmented  Data Science]{Augmented Data Science\\ Towards Industrialization and Democratization of Data Science}

%

\author{Huseyin Uzunalioglu, Jin Cao, Chitra Phadke, Gerald Lehmann, Ahmet Akyamac,  Ran He, Jeongran Lee, and Maria Able}
\orcid{1234-5678-9012}
\affiliation{%
  \institution{Nokia Bell Labs\\
  \{firstname.lastname\}@nokia-bell-labs.com}
}

%
\renewcommand{\shortauthors}{H. Uzunalioglu et al.}

%
\begin{abstract}
Conversion of raw data into insights and knowledge requires substantial amounts of effort from data scientists. Despite breathtaking advances in Machine Learning (ML) and Artificial Intelligence (AI), data scientists still spend a majority of their effort in understanding and then preparing the raw data for ML/AI. The effort is often manual and ad hoc, and requires some level of domain knowledge. The complexity of the effort increases dramatically when data diversity, both in form and context, increases. 
In this paper, we introduce our solution, Augmented Data Science (ADS), towards addressing this ``human bottleneck'' in creating value from diverse datasets. ADS is a data-driven approach and relies on statistics and ML to extract insights from any data set in a domain-agnostic way to facilitate the data science process. Key features of ADS are the replacement of rudimentary data exploration and processing steps with automation and the augmentation of data scientist judgement with automatically-generated insights. We present building blocks of our end-to-end solution and provide a case study to exemplify its capabilities.

\end{abstract}

%
%

\ccsdesc[500]{Information systems~Data mining}
\ccsdesc[300]{Mathematics of computing~Statistical paradigms}

%
\keywords{Machine Learning, Data Science, Augmented Intelligence}

%

%
\maketitle

\input{introduction}

\input{overview}

\input{SDD}
\input{feature}
\input{model}
\input{deploy}

\input{case}
\input{conclusion}

%
\bibliographystyle{ACM-Reference-Format}
\bibliography{main}

%
\section*{Acknowledgement}
We would like to thank our past summer students for their contributions related to this work: Rajarshi Bhowmik, Junzhuo Chen, Peng Wu, Yibo Zhao and Linjun Zhang (names in alphabetical order). We also would like to thank the anonymous reviewers for their constructive feedback on an earlier version of this paper.
\end{document}

%% file: introduction.tex
\section{Introduction}
\label{sec:introduction}

In typical data science projects, most of the data scientist's time, up to 80\%, is spent on pre-processing and feature engineering tasks~\cite{lohr2014, Caruana2015ICML}. These tasks prepare the raw data for the modeling tasks, where data mining and/or machine learning algorithms are applied. Interestingly, these tasks are often considered as art as well as science~\cite{peng_matsui_2016}, and are performed in an ad hoc manner by data scientists based on their knowledge and experience in the subject matter and in data science techniques. The need for the domain knowledge would require substantial interactions with domain experts. Often, the outcomes are lengthy data science projects that fall short of the expectations. 
Thus, even though we live in a data-rich world with access to ample computing resources and a large set of off-the-shelf algorithms, data science is experiencing what we call a `human bottleneck'. In this paper, we  introduce {\em Augmented Data Science (ADS)} towards removing this human bottleneck. Our goal is to {\em industrialize} and {\em democratize} data science.  Industrialization ensures that data scientists can extract insights from data in a much faster way without needing deep knowledge of the data domain. Democratization enables domain experts to perform basic data science tasks easily without being data scientists themselves.

The key insight behind ADS is that statistics and machine learning -- the technologies that data science facilitates -- are used to significantly remove the inefficiencies in various data science processes. Our goal is to help the practitioner, whether a data scientist or a domain expert, to convert raw data into insights at scale. 

ADS is an interactive, data-driven solution that augments the practitioner while performing knowledge extraction from any set of diverse data.
Our solution has the following key features:

\begin{itemize}
    \item {\em End-to-end:} ADS  solution targets all data science steps such as  data exploration, cleaning, feature learning and engineering, model building and maintenance.
    \item {\em Domain independence:} Techniques used in ADS do not assume the knowledge of problem domain, and thus are applicable to diverse problems and data sets.
    \item {\em Robustness:} No assumptions are made regarding the quality of the raw data. The system learns the data quality by itself and facilitates the practitioner's learning of it, potentially, in an interactive fashion.
    \item {\em Task-agnostic:} Steps prior to model building do not assume the knowledge of the downstream task, which can be supervised (e.g., classification) or unsupervised (e.g., clustering).  Certainly, knowledge of the task would help improve the early steps; however, our initial design objective is to create  generalized techniques that can be enhanced later with such knowledge. 
    \item {\em Augmentation:} ADS augments the user when automation is not feasible. This enables the user to make a better and faster judgement.
    \end{itemize}

%% file: overview.tex
\section{Previous Works}

Recent years have witnessed a diverse set of approaches toward democratization of data science. In this section, we provide an overview of approaches that address the end-to-end data science process as opposed to approaches that target only a subset of this process. The latter approaches are described in the relevant sections of this paper. 


Graphical tools~\cite{knime, rapidminer} have been developed to simplify the formation of the end-to-end data science process and create solutions without writing a single line of code. However, users of these tools 
still need to 
manually go through the entire pipeline.  
More recently, {\em Automated Machine Learning (AutoML)}~\cite{Caruana2015ICML} has been proposed to 
 implement algorithmic approaches towards democratization. A common approach towards AutoML is the representation of the modeling choices that a data scientist would make and then use an efficient search technique to optimize the model. 
For example, auto-sklearn~\cite{NIPS2015_5872} combines 4 data pre-processing and 14 feature pre-processing methods with 15 classifiers to create a structured hypothesis space with 110 hyper-parameters, and uses Bayesian Optimization 
to find the best model for supervised learning. Similarly, Tree-based Pipeline Optimization Tool (TPOT)~\cite{Olson2016} addresses AutoML
using genetic programming. However, these techniques assume data exploration and clean up have been performed by the user, and thus focus on automating the remaining steps of the pipeline.
More recently, solutions based on Deep Neural Networks (DNNs)  have been proposed \cite{LeCunBH15}.
However, finding the best architecture and tuning its hyper-parameters are non-trivial and time consuming.
Techniques to automate this process have been proposed \cite{NAS_autoML}; however, they are
computationally expensive as they train large numbers of DNNs.
Furthermore, DNNs are not applicable to small data sets or when data understanding is required.



TensorFlow Extended (TFX) was presented in 
\cite{TFX2017} 
as a general-purpose machine learning platform.
Although TFX aims at the end-to-end data science process, similar to the solution presented in this paper, its main focus is the discovery of data change and the update of the deployed models. Our solution, on the other hand, focuses on extraction of insights from data completely new to the users, and hence the solutions are complementary to each other.

There are also commercial tools such as DataRobot~\cite{datarobot}, H20.ai~\cite{h2o}, and SAS~\cite{sas} that focus on the end-to-end pipeline similar to the solution presented in this paper. Unfortunately, it is challenging to compare and contrast the techniques used in these products to our techniques as publicly available information do not provide details of the algorithms used. However, availability of these products proves the importance of automation and augmentation of data science for the industry,

\section{Solution Outline}
The first step in a typical data science project is the translation of a need or question into a data problem. Based on the problem statement, proper data is collected for exploration. Initial data exploration includes steps to correct data quality issues (re-collection, pre-processing to remove outliers, imputation to compute missing data, etc.). The data scientist uses domain knowledge to explore and manipulate the data, as well as to perform feature engineering, a process to create data representations that are more insightful about the problem being investigated. Finally, a model (classification, regression, clustering, etc.) can be built from the resulting data set. Although we describe this process as sequential, the process is iterative where the data scientist can revisit the earlier stages based on the insights gathered at any stage. Once the modeling results are satisfactory, the model can be deployed into an operational system for application to new data. To ensure that the model performs as expected in the deployed system, its performance needs to be continuously monitored. The model may need to be re-built if the target system varies or the model performance changes over time. Note that many variations of this process are possible; however, for the clarity of presentation, we describe our contributions in this paper along with this basic end-to-end flow as three separate modules.


\subsection{Augmented Data Discovery and Pre-processing}
As each data set has its own context, characteristics and quality, the data scientist would need to develop an understanding of the data, often in an iterative way, using basic statistical tools and techniques and visualisation. Each performed task adds to the understanding of the data; however, there is no single recipe that can handle every data set. Instead, a recipe is created by the data scientist as the learning of the data is progressing. As such, data discovery and pre-processing require laborious efforts and creativity from the data scientist. Our solution introduces a {\em Smart Data Discovery (SDD)} module that eases the data scientist's burden by automatically generating  insights about the structure and the quality of the data. These insights eliminate many rudimentary tasks that the data scientist would otherwise need to perform, as well as provide information about the data to the user and the downstream tasks such as feature engineering, which can utilize it for automatically constructing informative features. The details of the SDD module are provided in Section~\ref{sec:sdd}.

\subsection{Automated Feature Engineering}
The goal of feature engineering is to transform the raw data into a form that is more amenable to downstream analytics and/or machine learning tasks. This transformation can involve changing the shape of the data (e.g., from event records and temporal form into tabular form) as well as enriching the data contextually (e.g., computing statistical summaries of temporal data). This task generally requires subject matter expertise for the problem and the data set at hand. Thus, data scientists, if they are not already subject matter experts (SMEs), would need to learn from the SMEs about the problem space to generate new features. As this is often an iterative and time-consuming process, it represents a great opportunity for augmented intelligence algorithms to help the data scientist. Our solution automatically generates new features based on the knowledge gained from the SDD module without relying on any domain information. Data scientists may still generate new features through traditional means and with the aid of insights SDD module provides, however, automatically-generated features can prove to be sufficient in many cases. The details of automated feature engineering are provided in Section~\ref{sec:feature}.

\begin{figure*}[!t]
\centering
\includegraphics[width=0.8\textwidth]{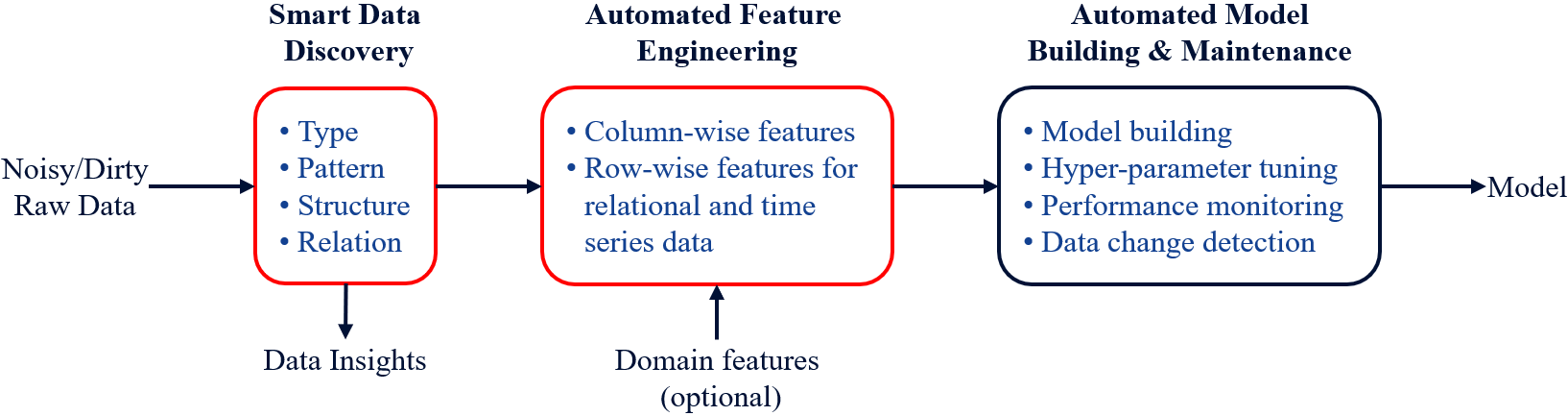}
\vspace*{-0.1in}
\caption{Diagram showing modules of ADS solution with their main functions}
\label{fig:E2Eflow}
\end{figure*}

\subsection{Automated Model Building and Maintenance}
\label{ambintro}
Once a training data set is prepared, the next step is to construct a model that provides an answer to the original data question. As many off-the-shelf algorithms are available for this task, the goal of the data scientist is to identify the best model with its hyper-parameters. In general, this task is formulated as an optimization task and performed through grid-search or more efficient search schemes~\cite{NIPS2015_5872, Olson2016, Bergstra.Bengio:2012}. Although the optimization task does not require manual labor from the data scientist, better results can be achieved if the training data is manipulated specifically for different types of learning algorithms, which would require data scientist labour. Our Automated Model Builder (AMB) module handles this step automatically before performing model search and hyper-parameter tuning and is explained in Section~\ref{sec:model}.

Once a model is deployed into production, the performance of the model should be verified to ensure it generalizes well to unforeseen data. If the training data that was used for model building has the same statistical characteristics with the new data and best-practices  for model fitting  were followed, it can be expected for the deployed model to perform similar to its performance in the modeling step. However, the model performance may degrade with time if the characteristics of the data change. This can be easily detected with the drop in model performance, although this may not be the fastest detection mechanism for certain applications as the model performance may not be available immediately. Thus, in our solution, we developed a scalable data change detection mechanism, which is explained in Section~\ref{sec:model}.

\subsection{Summary}
Our solution is summarized in Figure \ref{fig:E2Eflow}. The solution has a number of critical aspects. First, it addresses the entire machine learning pipeline, starting from the initial data exploration and cleaning, which is probably the most time-consuming step. Second, we rely on augmentation of the user in addition to automation, as human judgment, augmented with statistics and machine learning, would avoid the explosion of the hyper-parameter space that would result from full automation approaches. Finally, our technique also facilitates data understanding and can be useful for unsupervised learning tasks as well.

In the following sections, we will describe the main functions of the modules comprising ADS solution. The main focus of our description will be on  first two modules  which is not adequately addressed in the AutoML literature.
In Section \ref{sec:sdd}, we describe augmented data
discovery and pre-processing, consisting of automated data discovery (Section \ref{subsec:metadata}) and automated structure discovery (Section \ref{subsec:datastruct}), as the first modules of our system. Section \ref{sec:sdd} concludes with a small illustrative example of augmented data discovery (Section \ref{subsec:example}). In
Sections \ref{sec:feature} and \ref{sec:model}, we describe automated feature engineering, and
automated model building, respectively. In Section \ref{sec:case}, we present a
case study of applying our system to the end-to-end exploration of
a real-world dataset from the Kaggle Competition.

%% file: SDD.tex
\section{Augmented Data Discovery and Pre-processing}
\label{sec:sdd}

Prior to machine learning activities, the data scientist has to develop an understanding of the data, both in terms of the data types and data structure (e.g., dependencies or redundancies between the
various data fields). For a relational database, the contextual and structural information are stored in metadata, or simply put, data about data. The metadata are typically expressed in standard schema form, which is often generated by the domain experts at the time of data creation. Maintaining metadata schema can become a daunting and error-prone task on any big data, where it can contain hundreds, even thousands of fields. 
Even if database schemas or labels are available and up-to-date, developing a thorough understanding of the data is an iterative process, and often requires lots of manual effort and domain knowledge. 
For example, an expert in wireless networks can easily decide which fields in wireless performance data tables are important for a particular analysis, which is something the data scientist has to extract through statistical processes. 

Our solution introduces a {\em Smart Data Discovery}
(SDD) module to automatically uncover the data types, structure, and relations within a given undocumented data set without the need for domain knowledge. This module eases the data scientist's burden and augments the understanding of the data, providing key information to downstream tasks. In the following, we discuss
some of the main components and describe our innovations in the SDD module. To this end, we limit the scope to tabular data
and present our work in the context of individual data tables. 
The tabular data can be linked to form a database. In the next section, we show how the discoveries from individual tables can be potentially combined to create more complex features. 
A key design goal of our methods is robustness against a small percentage of polluted data, termed as {\em dirty records}, as we do not have prior knowledge about the data quality in advance. In fact, a by-product of our solution is the identification of dirty records, which can be used to assess data sanity. 

Data in tabular form is organized by {\em rows} and {\em columns}, where each {\em row} is one instance of the type of object described in that table, and each {\em column} is 
a set of data values of a particular simple type, one value for each row of the table. In general, a column may contain text values, numbers, or even more complex data types such as whole documents, images or even video clips. 
A row can also be called a record, and a column can also be called an attribute. Broadly speaking, we consider two types of automated discovery: automated data discovery and automated structure discovery. 

\subsection{Automated Data Discovery}
\label{subsec:metadata}
Automated data discovery refers to the discovery of information related to a single column, and the preparation of this data. We describe four techniques here.

\subsubsection{Data Types}
  Data types can be characterized at several levels: the primitive type such as integer, float, text, etc.;  types that describe the measurement scales such as nominal, ordinal; types that give contextual understanding, such as a person's name, datetime, zip code, etc. Automated inference of the data type is considered as an open problem \cite{Caruana2015ICML}. 
  
  At the primitive level, there are many existing tools for type inference. For example,  parsers in Python or R infer the data types automatically by reading a sample of rows. However, these tools typically do not allow for any dirty records, hence the inference may change in the presence of dirty (e.g., inconsistent form of) data. 
  Furthermore, these tools derive the types based on the superficial appearance of data, rather than its ``semantic'' meaning. For example, an {\em employeeID} column may be numeric by its value, but it does not carry any numeric meaning and should be treated as nominal instead. 
  Our solution 
  identifies the type that is mostly consistent with \cite{avro}, while allowing for a small percentage of dirty entries. Intelligently determining the measurement scale of a column is one of the areas of our current research. 
  
  At the contextual level, for example in the case of tables with alphanumeric columns, unless the schema or column types are well defined, up-to-date and known, the data scientist has to manually assign contextual labels to these columns. This can be quite time-consuming due to the sheer number of fields, and thus the automation of context discovery is an important pre-processing step that augments the scientist's understanding of the data. The field of schema matching is well-studied  for numeric fields \cite{rahm.bernstein.schema}, however extracting the context of string or alphanumeric data is an open problem and a subject of our current research \cite{akyamac.context}.
  

 \subsubsection{Quantum}
 For a data column that is numeric, it is helpful to know 
 whether its values have a smallest measurement unit, or 
 {\em quantum}. This may be especially important for time series data if it is regularly collected, 
  so that the collection time interval can be automatically extracted while allowing for measurement errors. 
 In \cite{he.2018}, we developed a statistical approach to infer the quantum in the presence of data noise. Upon detection of such noise with data, the data scientist can explore the noisy samples and decide on a course of actions.
 
 \subsubsection{String Pattern Discovery}
 \label{subsec:string}
  Many text columns have certain formats or patterns, for example, a common format of the datetime column is `Year-Month-Day:Hour-Minute-Second'. Another example is email address which is usually `name@domain' and the domain could have {\em com}, {\em org}, {\em edu} suffixes. 
 These are examples where prior knowledge has already been obtained. 
 To identify whether a text column has a certain pattern, and then to extract important sub-column features that might be useful for a downstream learning task, {\em without using prior knowledge}, is an important problem. 
 Although text mining has received considerable attention in recent years, (see \cite{AllahyariPASTGK17a} for a recent survey), there is little research that focus on such inference. A regular expression based solution  was developed in~\cite{tdda} without a formal formulation and  no guarantee for the obtained patterns. We have developed a solution that can identify such pattern with the presence of dirty records \cite{cao.sdd}. We formulated the problem as finding the maximal common sequence $S$ for a collection of $N$ strings, ${\mathcal S} = \{S_1, S_2, \ldots, S_N\}$, up to a percentage value $p$, i.e., $S$ is a maximal common sequence for at least $Np$ of the strings in the collection ${\mathcal S}$. 

 \subsubsection{Enumeration of Text}
 Textual data may be encountered in unstructured (as free-form text) or semi-structured form (e.g., as machine-generated text or system logs). Often, a structure is not known or well-defined, or is difficult to infer using filtering or techniques based on regular expressions. If the collection of text across the corpus represents a limited set of topics or messages, it can be converted to a categorical representation to be used in lieu of the original text itself. The resulting text categories can then be used in downstream machine learning tasks as enumerated variables. We developed an automated approach that uses intra- and inter-document properties to categorize textual data into such enumerated variables \cite{akyamac.text.processing}.

\subsection{Automated Structure Discovery}
\label{subsec:datastruct}

Automated structure discovery refers to the discovery of information related to multiple columns. We describe four techniques related to functional and statistical dependencies, and the structure of missing data.
 \subsubsection{Candidate / Primary Key / Functional Dependence}
 \label{subsec:schema}
 A candidate key is a combination of attributes that can be uniquely used to identify a table record without referring to any other data. Each table may have one or more sets of candidate keys. One of these candidate keys is selected as the table primary key which is given in the schema. 
 Information on candidate and primary keys can greatly improve data understanding. 
 
 
 The set of all candidate keys can be computed from the set of {\em functional dependencies}. In simple words, the set of attributes $X$ functionally determines $Y$ if and only if
 each set of $X$ values is associated with precisely one $Y$ value. The concept can be relaxed to define {\em near functional dependency} to include a small percentage violation. 
Several algorithms have been proposed to derive the functional dependencies (see \cite{Liu:2012} for a review). 
 However, in the worst case, the number of functional dependencies grows exponentially in the number of columns, which implies that exhaustive derivations may be expensive. 
 However, for general machine learning purposes, it is not necessary to obtain all such functional dependencies. 
 Our work \cite{cao.schema.diagram} does not focus on exhaustive derivation of all functional dependencies, but to obtain a schema tree diagram to represent functional dependence structure for downstream feature engineering.
 
 \subsubsection{Robust Constraint Discovery for Numeric Attributes}
 Functional dependency is deterministic, and often involves nominal attributes. One form of such dependency between numeric attributes is linear constraint in the form: $a'X = 0$, where $a$ is the vector of coefficients, and $X=(X_1,\ldots, X_k)$ are the $k$ attributes in such relation. For example, the Unix time can be expressed as a linear combination of `year', `month', `day', `hour', `minute', and `second'. 
 
 Extracting such linear relationships from the data
 not only helps us understand the data better, but  may also help us eliminate redundant columns when the data is used as input to a machine learning algorithm. However, uncovering such relationships in a robust fashion (i.e., allowing for a small percentage of violations) is non-trivial. 
 In \cite{cao.sdd}, we developed an approach where we first perform robust subspace estimation, and then drive sparse constraints.

 \subsubsection{Probabilistic / Statistical Dependency}
 \label{subsec:corr}
 Functional dependency considers dependency between  attributes that are deterministic. A generalization of this is statistical (probabilistic) dependency between attributes $X$ and $Y$, formally defined as the distribution of $Y$ conditioning on $X$, given as $f(Y\ |\ X) = p_X(Y)$ depends on $X$. If $p_X(Y)$ is independent of $X$, then $X$ and $Y$ are independent. 
 
 Understanding pairwise dependence between attributes may be the first step in exploring such  dependence and various scores have been developed for this purpose depending on attribute types. For example, 
 {\em Spearman's Rank Correlation} is used to measure correlations between numeric attributes; 
 {\em Goodman-Kruskal score} 
 is used to measure dependency between nominal attributes. We use normalized mutual information to measure dependency between nominal and numeric attributes. 
 More complex graphical models \cite{Barber2012} can also be developed to represent dependency structure between multiple columns.  
 
 \subsubsection{Structure of Missing Data}\label{subsec:missing}
 
Often the missing data arises systematically. Blindly applying imputation techniques without understanding of the missingness may potentially create bias in model building \cite{Scheffer02}.
To better understand the missing data structure, we convert each column into a vector of 0s and 1s, where 1 indicates missing.
The pairwise correlation of these missing attributes can be obtained using methods discussed in Section \ref{subsec:corr}. We also developed a tool that can be used to extract systematic missingness relationships between attributes by carrying out the following steps:
\begin{enumerate}
\item Clustering: Cluster the features based on similarity of missingness patterns. 
\item Modeling: Uncover the structure of missingness using predictive model on the cluster of interest. 
\end{enumerate}
The tool interactively finds the contextual patterns of missingness. 
The user can then decide to use that information for data cleaning or feature engineering. This is an instance of augmented data science in addition to automation.

\subsection{Illustrative Example}
\label{subsec:example}
We illustrate some steps of {\em Smart Data Discovery} using an example consisting of three relational tables, where a small sample of records are shown in Tables \ref{tab:customer}, \ref{tab:product} and \ref{tab:order}.
The three tables store information from customer purchase orders on an e-commerce website: a {\em Customer} table with information on the customers, a {\em Product} table with information on the e-commerce products, and an {\em Order } table with information on the actual customer purchases. The three tables are linked by the foreign keys {\em customerID} and {\em produceID} on the {\em Order} table. Contextual meanings of the columns in each table are self-explanatory here, however, in real-world scenarios this information may be beyond the data scientist's subject matter expertise. Note that the three tables contain missing and dirty entries.


\begin{table}[!t]
\caption{A Small Sample with Dirty Entries from Customer Table}
\label{tab:customer}
\vspace{-3mm}
{\small
\centering
\begin{tabular}{c| c| c | c | c}
 \hline\hline
    customerID & email & fullname & phoneno & age\\ \hline
    A1 & jw@gmail.com & John Smith & 908-544-2331 & 45\\
    2 & em@gmail.com & NA & NA & NA\\
    3 & cb@rutgers.edu & Emma Baker & 732-548-2331 & 40\\
    4 & as@yahoo.com & NA & NA & NA \\
\hline\hline
\end{tabular}
}
\vspace{-2mm}
\end{table}

\begin{table}[!t]
\caption{A Small Sample with Dirty Entries from Product Table}
\label{tab:product}
\vspace{-3mm}
{\small
\centering
\begin{tabular}{c| c| c | c | c | c}
 \hline\hline
    productID & pname & ptype & price & weight & shippingcost\\ \hline
    1 & StSqD & book & \$15 & 0.5lb & \$4.5\\
    2 & iAkmw & clothing & \ 32 & 0.6lb & \ 4.6\\
    3 & NudWI & clothing & \$32 & 0.2lb & \$4.2\\
    4 & wFztL & games & \$18 & NA & NA\\
    5 & VZedw & grocery & \ -999 & 1lb & \$5\\
    6 & JJGrA & music & \$15 & NA & NA\\
\hline\hline
\end{tabular}
}
\vspace{-2mm}
\end{table}

\begin{table}[!t]
\caption{A Small Sample from Order Table}
\vspace{-3mm}
\label{tab:order}
{\small
\centering
\begin{tabular}{c| c| c | c | c}
 \hline\hline
    orderID & orderType & productID & customerID & time\\ \hline
    1 & web & 1 & 4 & day 1\\
    1 & web & 6 & 4 & day 1\\
    2 & web & 3 & 2 & day 1\\
    3 & web & 2 & 2 & day 2\\
    3 & web & 4 & 2 & day 2\\
    3 & web & 6 & 2 & day 2\\
    4 & phone & 5 & 3 & day 2\\
    4 & phone & 7 & 3 & day 2\\
\hline\hline
\end{tabular}
}
\vspace{-2mm}
\end{table}


The small samples were chosen for illustrative purposes; our aim here is to show  the desired outcome of this automated analysis, and how this information can be  used for downstream learning tasks.  

\begin{table}[!t]
\caption{Derived Primitive Data Types of the Customer Data}
    \label{tab:meta}
    \vspace{-3mm}
    \centering
    {\small 
    \begin{tabular}{c||c|c|c|c|c}
    \hline\hline
      COLUMN &  customerID & email & fullname & phoneno & age \\\hline
      TYPE &   integer & string & string & string & 
         integer \\
     PATTERN & 1 & *@*.* & * * & *-*-* & 1 \\\hline
    CONTEXT & ID & email & name & phone & NA\\\hline\hline
     \end{tabular}
     }
     \vspace{-2mm}
\end{table}

\subsubsection{Data Types, Quantum and String Pattern}
Table~\ref{tab:meta} shows the derived primitive data types and their patterns for the columns of the {\em Customer} data in Table~\ref{tab:customer}.
In our implementation we used a 2-level structure for the primitive data type. The top level consists of {\em integer, numeric}  and {\em string}. The second level contains a further breakdown with types that is mostly consistent with \cite{avro}: {\em Boolean/byte/short/int/long} for {\em integer}, {\em float/double} for {\em numeric}, and a special {\em datetime} sub-type for {\em string}. Our data type allows for a small percentage of dirty records (as configurable parameters).  

For a field that is either {\em integer} or {\em numeric}, the pattern value indicates the identified measurement unit {\em Quantum}. For example, columns {\em customerID} and {\em age} is identified as integer with quantum value of 1. Notice that there is a dirty entry for 1st row in {\em customerID} and {\em age} has a quantum value 1, identified based on all records not just records in the sample table.
For a field that is of {\em string} type, the pattern displays the inferred textual pattern, where the
common sub-strings are expressed, and the deviations are shown using the wildcard symbol `*'. The wildcard symbols can be used to extract features, as described below. The last row of Table~\ref{tab:meta} shows the desired outcome for the derived contextual types of columns.
Automatic discovery of this contextual information without using column names is a subject of our current research. 


We can then extract features for downstream feature engineering. For example,  the email column in Table \ref{tab:customer} can be mapped to three columns representing each of the wildcard values in the discovered pattern; the {\em price, weight} and  {\em shippingcost} in {\em Product} table  (Table \ref{tab:product}) will be reshaped by removing the suffixes and prefixes, respectively, and retrieving only the numeric content of the column. 

\subsubsection{Schema Diagram}
\begin{figure*}[!t]
{\small
\begin{minipage}{1\textwidth}
\centering
\subfigure[Schema Graph for {\em Order} Table]{
\includegraphics[width=.18\textwidth]{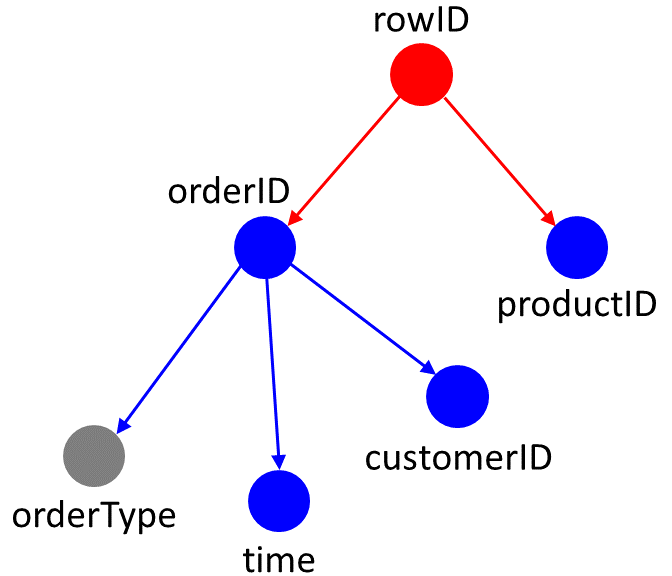}
\label{fig:OrderSchema}
}
\subfigure[Association Graph for {\em Product} Table]{
\includegraphics[width=0.22\textwidth]{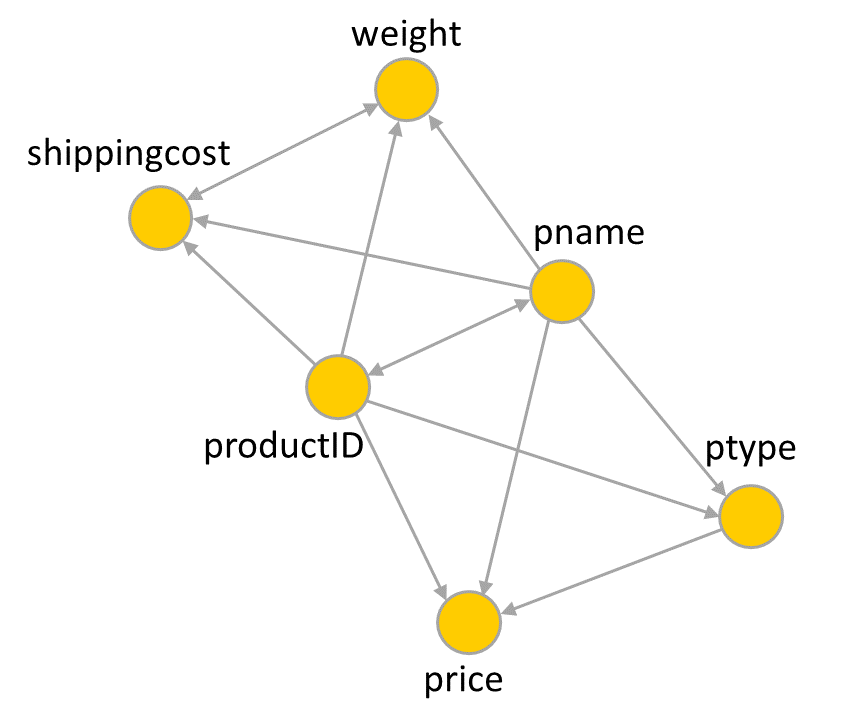}
\label{fig:corr}
}
\subfigure[Relation Graph for {\em Order} Table]{
\includegraphics[width=.2\textwidth]{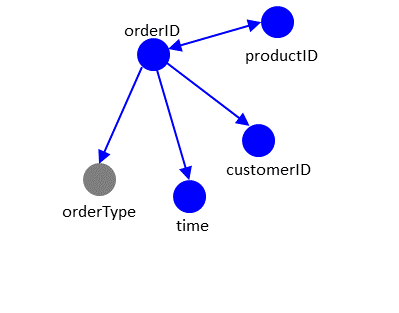}
\label{fig:OrderRelation}
}
\subfigure[Combined Relation Graph for all tables.]{
\includegraphics[width=.23\textwidth]{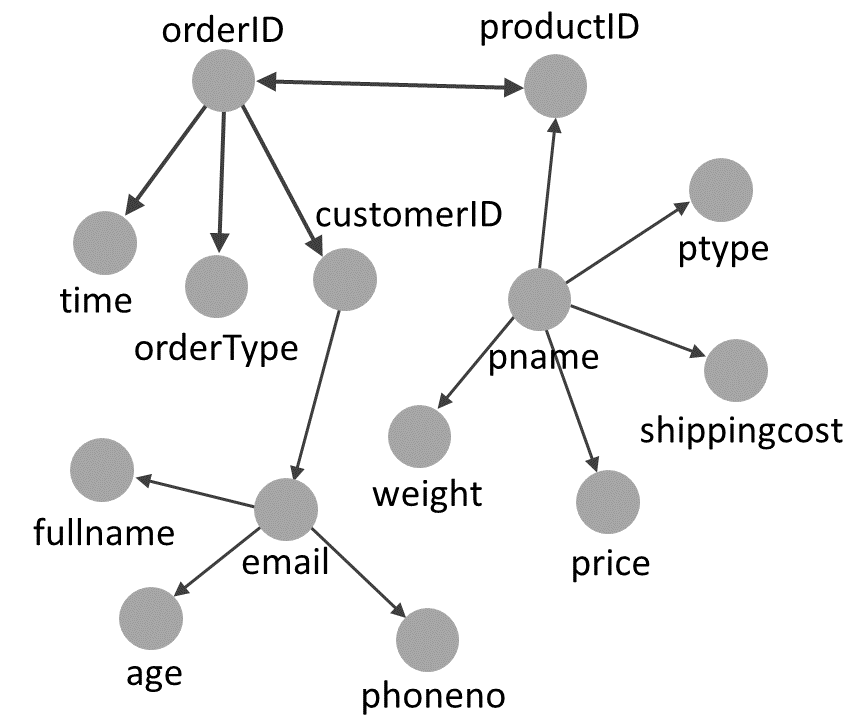}
\label{fig:combinedRelGraph}
}
\vspace{-3mm}
\captionof{figure}{Schema and Relationship Graphs for the Illustrative Example}
\end{minipage}
}
\end{figure*}

Figure~\ref{fig:OrderSchema} shows the discovered schema for the {\em Order} table (Table~\ref{tab:order}), using methods proposed in \cite{cao.schema.diagram}. 
Let $\rightarrow$ indicate a functional dependency, i.e., the set of attributes $X$ functionally determines $Y$ if $X\rightarrow Y$. Let $\leftrightarrow$ indicate the functional dependency in both directions for columns $X$ and $Y$, i.e., $X\leftrightarrow Y$ if and only if $X\rightarrow Y$ and $Y\rightarrow X$. In the diagram, blue nodes indicate the bidirectional functional dependency $\leftrightarrow$, and gray nodes indicate one directional functional dependency $\rightarrow$. Figure~\ref{fig:OrderSchema} can be interpreted as follows: 
The root node {\em rowID} (shown in red), indicates a hypothetical {\em rowID} column whose values are the row indices starting from 1. Then, 
\vspace*{-0.05in}
\begin{eqnarray}
rowID & \leftrightarrow & (orderID,
\ productID) \nonumber  \\
orderID & \leftrightarrow & (customerID, \ time) \nonumber\\
orderID & \rightarrow & orderType. \nonumber
\end{eqnarray}
In Section~\ref{subsec:path}, we show how we use these schema diagrams to define relational graphs and how we merge these relational graphs from multiple tables for feature engineering. 
In \cite{cao.schema.diagram}, we also illustrate our methods for finding the schema diagram using more complex datasets. 

\subsubsection{Statistical Dependency and Linear Constraints Discovery}
Figure~\ref{fig:corr} shows the association graph between nominal columns in the {\em Product} table (Table~\ref{tab:product}), based on 
the {\em Goodman-Kruskal (GK) scores}. 
An edge between  two nodes indicates a high association  between the two columns (we used 0.9 as the threshold). Unlike other association scores, $GK$ score between two attributes is asymmetric as it measures the extent to which variation in one attribute can be explained by the other.
For a pair of columns $A$ and $B$, let
$A\rightsquigarrow B$ denote high association from $A$ to $B$ by the $GK$-score, and $A\leftrightsquigarrow B$ denote high association between $A$ and $B$ in both directions. 
Notice that both $productID$ and $pname$ are equivalent keys for the table, it is obvious that the $GK$ score from these two columns to any other column is 1. Removing directed links from these two nodes in Figure~\ref{fig:corr} reveals additional important associations:
$ weight \leftrightsquigarrow shippingcost$ and $ptype \rightsquigarrow price$. These associations can be easily explained by the nature of the columns using domain knowledge; we automatically reveal these associations using our smart data discovery without requiring this knowledge. This will be especially useful when the data scientist needs to work with complex tables from arbitrary domains.

\subsubsection{Structure of Missing Data}
For brevity, we only present the results of the discovered missing data structure. 
There are two conclusions that we can reach from the analysis on how the missingness of the columns are related to each other: 1) In {\em Customer} table, columns {\em fullname, phoneno, age} are missing together.  2) In the {\em Product} table, columns
{\em weight} and {\em shippingcost} are missing together. In terms of a predictive model of the missing data structure, we conclude that for {\em Product} Table, {\em weight} and {\em shippingcost} are missing whenever the product is of type {\em games} or {\em music} (as expected). In this case, the missingness is systematic, and employing imputation techniques to fill in missing {\em weight} and {\em shippingcost} is not appropriate. In fact, 
for systematic missingness, it is probably important to create a missing label for downstream learning tasks.

%% file: feature.tex
\section{Automated Feature Engineering}
\label{sec:feature}

Feature engineering is an important task in any type of data modeling as it transforms raw data into features that are more meaningful to the analysis. Traditionally, feature engineering requires the knowledge of the semantics of the data, thus domain knowledge. Instead, we focus on the generation of features automatically. 

For tabular data, prior methods for automatic feature engineering can be classified into the following two categories. For the first category, new features are constructed {\em column-wise}, by applying a transformation function to one or several
columns (raw features in the data table).  For example, we can transform a nominal column using one-hot encoding, a numeric column by quantile binning,
or define a ratio by division of two columns.
More sophisticated transformations have been discussed in
\cite{Li2012ExtendingAI}, for example,
applying to those columns with detected dependencies such as those discussed in Section \ref{subsec:corr}.
For the second type, 
the features are constructed {\em row-wise}, by summarizing multiple rows that corresponds to the same entity of interest. For example, a time series of an entity will have multiple records ordered in time, and observed values through time need to be aggregated for that given entity. Automated feature aggregation have also been proposed for relational tables \cite{OneBM,DFS-MIT}. 

Our innovation concerns feature engineering of the second type, i.e., through row aggregation. 
Our approach differs from previous approaches \cite{OneBM,DFS-MIT} in that it discovers the relationship between a raw feature to be aggregated and the entity of interest 
from within the data, not from the given data schema. In addition, it also finds paths for aggregation from the discovered structure knowledge to 
generate features.




Features are computed for a specific entity of interest, which we term the {\em anchor}. The anchor is task-dependant and specified by the user. For the sample illustrative data described in Section \ref{subsec:example}, we may want to cluster customers by their purchase behaviours, in which case the unit for feature generation is the customer and the anchor is {\em customerID} or we may want to predict if an order will be fulfilled on time in which case the anchor is {\em orderID}. The feature generation process is flexible in the specification of the anchor and can easily generate features for any meaningful anchor specified from the dataset by the user.
We describe some important aspects of our methodology in the following sections. 

\subsection{Relationship Discovery}
To generate features, the type of aggregation and transformation to be done depends on the nature of relationship between the anchor column and the other columns in the dataset. 
Humans rely on either their domain knowledge or provided complex ER diagrams to uncover such relationships. On the other hand, we take advantage of the automatically-generated schema (Section \ref{subsec:schema}) to compute features as explained below. 


\subsubsection{Schema to Path Generation}
\label{subsec:path}
From the schema graph discovered in \ref{subsec:schema}, using the information of the candidate keys and the functional dependencies, we construct a ``relation'' graph that provides a representation of relationships between data columns. A relation graph consists of nodes, which represent all the columns in the dataset and the links between them, which indicate the type of relationships. An outgoing link represents a one-to-one relationship, while an incoming link indicates a one-to-many relationship. The relation graph also preserves the path of traversal in the data along the path. Note that the construction of the relation graph is independent of the analytic task at hand. However, the features to be generated are based on the anchor. Given  a relation graph and an anchor, we  compute paths from the anchor to each of the nodes in the graph and also the relationship along the path.

To illustrate the concept, consider the schema graph for the {\em Order} table in our illustrative dataset shown in Figure \ref{fig:OrderSchema}. The combined set of all blue nodes at level one from the top virtual node, $rowID$,  together represents the candidate primary keys for the table. Therefore, \textit{orderID} and \textit{productID} together are the combined keys for this table. Being joint attributes, the relationship between them is one-to-many, i.e., a single \textit{orderID} value can have many \textit{productID} values and vice versa. The attributes of \textit{orderID} are \textit{orderType}, \textit{customerID} and \textit{time}. These are one-to-one relationships. We transform this schema graph into the relation graph shown in Figure \ref{fig:OrderRelation}.  If our anchor is {\em orderID}, there exists three paths that have one-to-one relationships, i.e. $orderID$  $\rightarrow$ $orderType$, $orderID$ $\rightarrow$ $customerID$ and $orderID$ $\rightarrow$ $time$ and one path that has one-to-many relationship,  $orderID$ $\rightarrow$ $productID$. The appropriate data transformations  to generate new features are described in the following section. 

The extension of our approach to when data resides in multiple tables is fairly straightforward. We first discover the individual schema graphs for each table and transform them to the relation graph. Assuming that columns that are equivalent across tables also share the same name, we  merge the individual relation graphs to get a combined graph for the entire dataset and follow the  graph traversal to discover the paths from an anchor to all other columns in the full dataset. For our illustrative example dataset, the merged relation graph is shown in Figure \ref{fig:combinedRelGraph}.

\subsection{Feature Generation} \label{subsec:data.transform}
New features are generated based on the data type of the columns and the relationships along the path. 

\subsubsection{One-to-One Relationship} When the relationship between the anchor and the data column is one-to-one, the column is taken as is. We can also apply (column-wise) transformations to construct more meaningful features. For example, if the data type is  datetime, we extract the components of the date such as  day, month, year, etc.; if the type is string, we  extract 
sub-parts of a string with distinct patterns, discovered  as described in Section ~\ref{subsec:string}. 

\subsubsection{One-to-Many Relationship} When the relationship between two adjacent nodes on the path is one-to-many, we employ aggregation strategies to generate additional features. This is done by grouping and computing aggregates for the values on the column under consideration. For example, if the path is $A$ $\rightarrow$ $B$ and there is a one to many relationship from $A$ to $B$, then we compute aggregates on values of $B$, grouped by $A$. The aggregation function to employ is based on the data type of column $B$.  If the data type is numeric, we compute the standard statistical measures such as minimum, maximum, mean, standard deviation, count and sum.  For data type that is not numeric we provide a count and distinct count of the values. Additional domain-specific aggregation functions can also be used. 

\subsubsection{Time Series Data}

There exists a special case of the one-to-many relationship that is ordered sequence of data or a time series. The column that denotes the time order can be specified by the user, but can also be automatically discovered based on the data type. In such a case, a special time series treatment can be applied to the data measurement. The goal here is to generate an efficient feature set capturing all kinds of temporal characteristics as an aggregation.
We developed a Time Series Processor (TSP) module based on previous work \cite{TSFRESH} that provides a broad set of feature generation functions for numerically represented signals, enriched by further functions found to be useful. 
A variety of features are generated  based on time-based summarizations (in terms of ranges, peaks, symmetries, etc.), transformations (to the frequency domain e.g., continuous wavelet transformation), autocorrelations, linear and non-linear behavior, sample distributions, and etc. The TSP also supports the aggregation of categorical indications over time.


\subsection{Feature Roll-up} \label{subsec:data.rollup}
One of the strategies  we employ during feature aggregation of data along a path is {\em feature roll-up}. A path may consist of a chaining of many one-to-many relationships. In such cases, one needs to roll-up the newly generated features from the end of the path to the anchor. For example, consider the path from a customer to price of the product, i.e., $customerID$ $\rightarrow$ $orderID$ $\rightarrow$ $productID$ $\rightarrow$ $pname$ $\rightarrow$ $price$. There is a one-to-many relationship from \textit{customerID}  to \textit{orderID} , and from \textit{orderID} to \textit{productID}. Features for the $customerID$ that are generated based on this path start to aggregate at the lowest level and begin to roll-up towards the anchor. So, first one would compute aggregated features for the $price$ per (\textit{customerId, orderId}). Then each of these generated features would then further be aggregated per \textit{customerID} to get the final set of features. 

Our techniques for automated feature generation may lead to a feature explosion. It is possible that some of the features generated are not meaningful and may not help towards the modeling. Feature selection strategies  employed just before the modeling helps alleviate this issue. More details of our approach to automated feature engineering can be found in \cite{phadke.auto.feature}. We recognise the fact that certain features are very domain specific and rely on human experts for their generation. Our goal in automated feature engineering is not to completely replace the human, but to augment the data scientist in generating as many features as possible in an automated fashion. 

%% file: model.tex
\section{Automated Model Building and Maintenance}  \label{sec:model}

As discussed in Section \ref{ambintro}, much of the work on AutoML focuses on automated model building after
the data exploration is performed. 
The model building process involves data pre-processing steps such as handling of missing values and splitting of the data set into training and test sets, in addition to
feature selection, transformation and combination, and selection of model hyper-parameters. Grid search and other optimization techniques can be
employed to select features and best parameters.
However, this task can become computationally prohibitive, particularly
without domain knowledge or data scientist input. Better results can be achieved if data manipulation and feature selection can be tailored to the machine learning 
algorithms being used. Our Automated Model Builder (AMB) automates the process of data manipulation and feature selection, in addition
to pre-processing steps like handling of missing values and splitting into training and test sets.  
AMB employs a greedy feature selection cross-validation loop that searches for the gain of feature combinations in concert with
a restricted feature pre-selection process using pairwise statistics and statistics combined with label information. These are performed in conjunction with parameter searches taking random search \cite{Bergstra.Bengio:2012} as a baseline using several machine learning algorithms.

%% file: deploy.tex

Model deployment is an important step in the machine learning pipeline where a previously established model is used for prediction on unseen data. Although changes in data characteristics do not necessarily lead to model performance degradation, the reverse is true, i.e., if the model performance degrades, then there must be some changes in the data distribution. 
Therefore to maintain model performance and understand the root cause of the degradation, a critical aspect here is to monitor and understand changes in data so that follow up actions would be triggered.
What makes the problem more challenging is that often the data are high-dimensional and the possible changes can occur in many different directions.

In \cite{datadiff}, authors proposed a solution, analogous to ``text diff'',  for comparing the repeated data
samples against each other. Their solution is referred as ``data diff'', 
which aims to find a ``patch'' to  transform the data samples so that
they are identically distributed. 
In \cite{zhao2019}, we proposed a solution
by using a non-parametric method for testing the equality of two multivariate distributions for high-dimensional data. Our method is based on 1-dimensional random projections and the classical Kolmogorov-Smirnov (\textit{KS}) test. Our proposed method is distribution-free, i.e., with no assumptions about the forms of the distribution, 
and scalable for large sample size and data dimension.
At the same time, it provides information on where the changes have occurred.

%% file: case.tex
\section{Case Study}
\label{sec:case}


The effectiveness of our techniques can be assessed in two dimensions: 1) The time and effort saved when compared to a manually performed analysis done by a skilled data scientist using tools he or she is experienced with,  and 2) the measurable performance of the results with respect to the data analysis target (e.g. the classification performance in case of a classification problem). The former requires a carefully designed user study, which we plan as future work. In this section, we address the latter by applying ADS into a real-world data set from the Kaggle Competitions to predict Home Credit Default Risk \cite{kaggle}. For this example, we present automatically-generated data discovery results and compare the competition task performance with that of the competition winner. While we acknowledge that a fully automated technique may not be able to beat a human expert out of the box, it can come very close to it while saving a tremendous amount of effort. The details of our study are presented below.

\subsection{Data Description}
Home Credit provides loans to consumers with insufficient credit histories. Data about applicants, their previous loan and payment histories as well as  credit card transaction histories were provided. The aim was to predict if a particular
 user would default on his/her loan. The data consisted of a total of 7 tables: a {\em Main} table with key client attributes, two tables for clients with previous loan applications from other institutions: {\em Bureau} and {\em Bureau.balance} and 4 more tables describing application data for clients with previous loans
in credit: {\em Previous.Applications, POS.Cash.Balance, Install.Payments} and {\em Credit.Balance}. Tables and the columns (keys) that connect two tables is provided. Our aim was to discover the data structure and relationships within the data and use that to generate features and then model the prediction, a supervised binary classification problem.  The dataset has  nearly 308K applicants in the training set and 48K applicants in the test set. There are over 50 million combined transactions for all the applicants and each transaction has several attributes (average 20). The total data size is over 2.5GB.

\subsection{Data Discovery}
\subsubsection{Data Types, Quantum and String Patterns}
Most tables have fields with a mixture of numeric and string type. The primitive data types appear to be mostly clean, although there are some non-conforming ones. For example, {\em $AMT\_GOOD\_PRICE$} in the {\em Main} table is identified as type {\em integer}, with less than 0.2\% non-conforming values (these are decimal valued instead of integers). However, these anomalies appear to be inconsequential. Many numeric columns are integer valued, and for most of these, the detected quantum is 1. There are a few exceptions, for example, {\em $AMT\_GOOD\_PRICE$} has an identified quantum value 4500, {\em $AMT\_INCOME\_TOTAL$} appear in increments of 250.
Strings in these tables typically have a few distinct values. There is no significant pattern found that is useful for data shaping.

\subsubsection{Schema Diagram}
We run our tool to derive the schema diagram, and cross-checked with the information given on the competition website. All tables are in agreement except for the {\em Installment.Payment} table. It appears that the primary key for the table was not completely described in the online documentation. Other than the information that was given, we also identified some near functional dependencies, for example, in the {\em Bureau.balance} table, columns {\em SK\_ID\_CURR, DAYS\_CREDIT\_UPDATE}, {\em DAYS\_CREDIT} {\em\_ENDDATE} is almost a row identifier, except for about 1\% of rows; column {\em AMT\_CREDIT\_SUM\_LIMIT} is functionally dependent on columns {\em SK\_ID\_CURR} and {\em DAYS\_CREDIT\_UPDATE}. 
However, currently in our feature engineering, we do not explicitly use these near functional dependencies to design aggregated features. 
This is an area of our future work.

\subsubsection{Association and Dependency}
\begin{figure}[!t]
\centering
\includegraphics[width=0.23\textwidth]{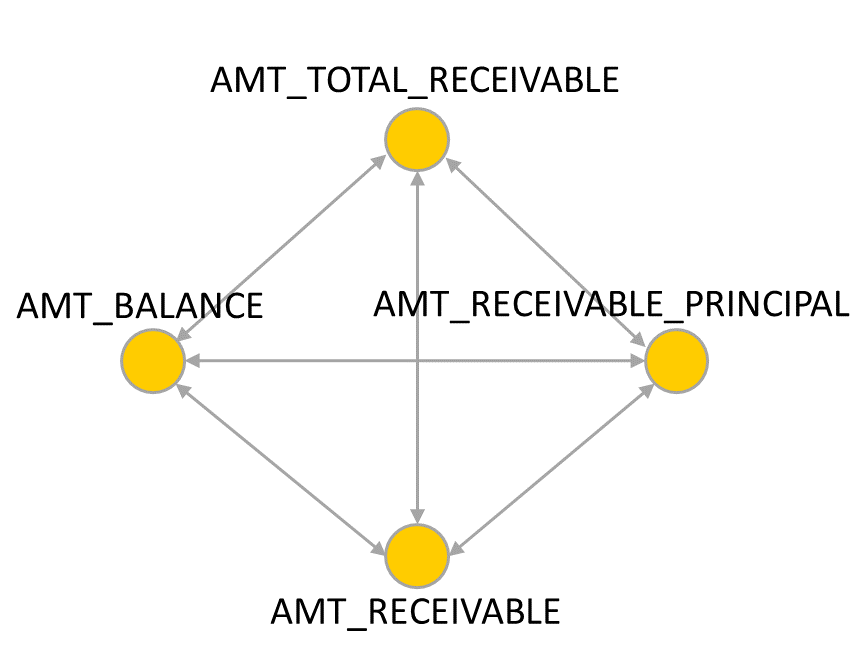}
\vspace{-2mm}
\caption{A clique showing highly correlated pairs of columns in the Association Graph for {\em Credit.Balance} Table}
\label{fig:creditCredit}
\vspace*{-0.1in}
\end{figure}

We found highly correlated columns in the database tables. Figure~\ref{fig:creditCredit} shows a cluster of four highly correlated columns in {\em Credit.Balance} table, with their ranked Pearson correlation over 0.97. Several such clusters are found in this table as well as in most other tables. In fact, in the {\em Main} table there is a large cluster of 55 highly correlated columns. As the features in the {\em Main} table are directly associated with clients and there is no aggregation involved, sophisticated model builders such as {\em lightGBM} \cite{NIPS2017_6907} and AMB (described in Section~\ref{sec:model}) will take those correlations into account to avoid overfitting. However, for columns not in the main table, since we need to engineer features via a roll up process (described in Section~\ref{subsec:data.rollup}), directly taking each column one by one without considering their correlation will likely to result in information loss due to the aggregation process. How to engineer features for these correlated columns as a whole is an area of our current research. 
We also note that no linear constraints were found except some trivial ones such as columns with dominant single values.

\subsubsection{Missing Data Exploration}
This dataset contains a lot of missing values, and many such missing values are systematic. We used our missing exploration tool discussed in Section~\ref{subsec:missing} for this analysis. We will demonstrate this using the {\em Main} table. 

\begin{figure}[!t]
\centering
\includegraphics[width=0.5\textwidth]{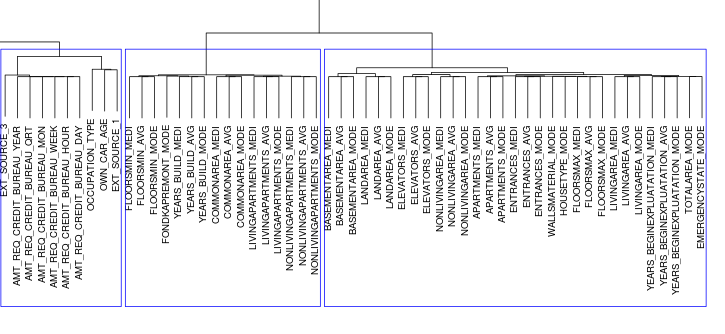}
\vspace{-8mm}
\caption{Columns in {\em Main} table are missing together, forming three clusters of distinct missing percentages.}
\label{fig:main.missing}
\vspace*{-5mm}
\end{figure}

Figure~\ref{fig:main.missing} displayed how columns in {\em Main} table are missing together. Columns that have missing values can be roughly clustered into three groups: Cluster 1 consists of 10 columns with 26\% missing values, Cluster 2 consists of 16 columns with 68\% missing values, and Cluster 3 consists of 31 columns with 51\% missing values.
In many instances, columns in the clusters are missing together. For example, for Cluster 3, about 73\% of rows have all of the 31 columns missing. Next we run a predictive modeling step in the attempt to explain causes of these missingness, using a simple Classification and Regression Tree (CART) on the group missing label of the three clusters. Results from CART show that if {\em FLAG\_OWN\_CAR} is `N' (no), then the group missing label will be 1 with 100\% probability. This implies systematic missingness in the data. In this case, it is not appropriate to impute numbers for these columns, but rather the more appropriate action is to add the group missing label as a potential feature. 

Similar systematic missing structure have been observed for other tables in the dataset. However in some cases, it is not easy to explain the missingness from other columns in the table. In this case, it may be especially important to add the group missing label to the model since that information is not captured by other columns. 

\subsection{Model Building and Results}
Based on the data discovery, we generate features for the dataset using the automated methodology described in Section \ref{sec:feature}. The anchor in this case is the applicant viz: {\em SK\_ID\_CURR}. We generate  features as described in Section \ref{subsec:data.transform} and get a total of 1385 features.
The model with the best prediction is {\it lightGBM} with an AUC of 0.781. The top of the leaderboard had an AUC of 0.817. Although some may argue that a difference of 4.6\% is significant, it should be noted that the leaderboard position had several hundred submissions, a proxy for the effort that went into getting to the top. On the other hand, our submission was automated and had just a couple of submissions. Examination of the discussion by some of the top scorers suggested use of ensemble methods to  combine predictions from several different models, while our current implementation focuses on a single model. We plan to extend AMB with such ensemble methods in the future. Hand-crafted features based on domain expertise can be augmented to the auto generated features to further improve model performance. Thus, our end-to-end methodology requiring very little human effort provides a good initial base model with a performance that comes within close reach of the top. 

%% file: conclusion.tex
\section{Conclusions}
\label{sec:conclusion}


We presented Augmented Data Science, 
where we apply statistics and machine learning to
extract insights from any data set in a domain-agnostic way to facilitate the data science process.
Our solution introduced a set of modules that automatically uncovers the quality and structure of the data, generates new features and finally identifies and tunes the best model for the task. Each of the modules also augments the user with a better understanding of the data and allows the user to intervene to provide expert judgement. The chaining of these modules in an end-to-end fashion provides for a fast and automated way to work with new and varied datasets. We validated the outcome with a real-world dataset and showed that it can work well even without domain expertise. This is an important step towards the industrialization and democratization of the data science process. In the future, we plan to work on enhancing some of the techniques as discussed in the individual sections and will build a system available for use by all. Furthermore, we are planning to perform a user study to evaluate the time and effort saved through the use of the ADS solution. We will also extend the performance evaluation to additional real-life data sets and different machine learning tasks. Finally, whether the whole data science pipeline can be fully automated or not is an open research problem. Reducing the need for human cognition will increase the search space for the end-to-end pipeline optimization. The evaluation of this trade off between augmentation and complexity remains as future work.